\documentclass[11pt]{article}

\usepackage{amsmath}
\usepackage{tgpagella} 
\usepackage{tablefootnote}
\usepackage[dvips]{graphics}
\usepackage[utf8]{inputenc}
\usepackage{amsmath}
\usepackage{amsfonts}
\usepackage{amssymb}
\usepackage{graphicx}
\usepackage{fancyhdr} 
\usepackage{multirow}
\usepackage{wrapfig, framed}
\usepackage{algorithm}
\usepackage{algorithmic}
\usepackage{epsfig,subfigure,endnotes,color,url,paralist, multirow,float}
\usepackage[left=2cm,right=2cm,top=2cm,bottom=2cm]{geometry}
\usepackage[inline,shortlabels]{enumitem}
\usepackage{mathtools}
\usepackage{mathrsfs} 
\usepackage[table]{xcolor}
\definecolor{lightgray}{gray}{0.9}
\definecolor{LightCyan}{rgb}{0.88,1,1}

\usepackage{lipsum}
\usepackage{comment}
\usepackage{soul}
\usepackage{tikz}
\usepackage{xcolor}
\usepackage[format=plain,
            labelfont={bf,it},
            textfont=it]{caption}
\usepackage{xspace}
\usepackage{hyperref}
\usepackage{booktabs}

\usepackage{cite}
\usepackage{url}
\DeclareUnicodeCharacter{2082}{\₂{}}
\usepackage[compact]{titlesec}
\usepackage[rightcaption]{sidecap}
\usepackage{mhchem}

\usepackage{placeins}

\usepackage{graphicx} 
\graphicspath{ {images/} }

\begin{document}

\newpage
\pagenumbering{arabic}
\pagestyle{plain}
\setcounter{page}{1}
\setcounter{section}{0}


\begin{center}{{\bf \LARGE
Multiverse Computing CompactifAI : Accuracy and Consumption Analysis from a Compressed Llama 3.1 model}}
\end{center}

\begin{table}[!h]
\centering
\begin{tabular}{m{0.2\textwidth} m{0.2\textwidth} m{0.2\textwidth} m{0.2\textwidth}}
\centering Damien FOVET\\ \emph{SopraSteria Group}
& \centering Shashank CHAMOLI\\ \emph{SopraSteria Group} 
& \centering Sarah OURY\\ \emph{SopraSteria Group} 
& \centering Srishti SINGHAL\\ \emph{SopraSteria Group} 
\end{tabular}
\end{table}

\section{Abstract}

\paragraph{}
This study evaluates the performance of a compression method, called CompactifAI, developed by Multiverse Computing, applied to the large language model Llama 3.1 8B\cite{llama}. The evaluation focused on model efficiency (in terms of energy consumption) and accuracy using respectively the frameworks Codecarbon\cite{codecarbon} and Ragas\cite{ragas}. A comparison was performed between the model compressed with CompactifAI\cite{compactifai}\cite{compactifai2} and its full-size version. Our findings reveal that the compressed model using CompactifAI not only significantly reduced the computational resources but also maintained the model accuracy, making the model  more efficient, scalable and cost-effective. 
\par

\section{Introduction}

\paragraph{}
The exponential growth of Large Language Models (LLMs) and their increasing demand has been a major factor leading to an unparalleled surge in computational requirements, followed by significant increase in energy consumption, financial overhead and a race for performance. Due to their large specific infrastructure and widespread use, AI systems have a wide range of high environmental impacts. Reducing these impacts can be achieved through a number of actions: reasoning about uses and giving priority to AI projects with positive impacts, optimizing software (model, data, languages...) and infrastructures. In this study, we will focus on the model optimization which is only one part of the approach needed to counter the negative impact of AI on the environment.

\paragraph{}
In pursuit of trying to determine what impact compressed models could have on computational and energy reduction, SopraSteria sustAIn team decided to evaluate CompactifAI from Multiverse Computing, a new innovative approach that uses tensor networks with other techniques that reduces the amount of parameters. The objective was to test the compressed version of Llama 3.1 8B compressed by CompactifAI against the original version and benchmark cost savings, energy consumption and model precision.

\paragraph{}
SopraSteria created a sustAIn team dedicated to frugal AI within its AI program called rAIse. This team is working in particular on AI software efficiency for the past few years. We have been studying the various factors influencing the energy consumption of an AI system based on a language model (small and large), but also from more simple AI methods such as regression or classification algorithms. We performed many tests, varying the language model, its number of parameters, the quantization, the framework... The aim is to help presales, developers, and project managers become aware of the impact of AI so they can make informed decisions. With these benchmarks, we are able to make impact predictions and provide them quantified best practices for more frugal AI so they can implement eco-design from the earliest stages of the project. To reduce the model's size and therefore its energy consumption, the team had already studied classical model compression methods such as quantization. Their gains in terms of energy consumption had been proven, with reductions in the order of -20\%, according to our experiments. However, we can see that there are limits to using this type of method, as it can lead to large losses in accuracy, which is not always acceptable for some systems. When we heard about this innovative compression method from CompactifAI, we wanted to check that the efficiency gains were significant and that the loss of accuracy was low enough to promote its use in our projects.  

\hfill

\paragraph{}
In the next sections, we describe how we performed our experiments to collect data and explain the results we obtained. Then, we provide a critical discussion about the limitations of our work and the additional work that may be required for more complete conclusions.

\section{Methodology}

\subsection{Experimental setup}

\subsubsection{Experimentations description}

\paragraph{}
We have adopted a dual-faceted analytical approach to systematically assess the effects of model compression. It includes:

\begin{enumerate}
\item Power Consumption Analysis: A quantitative evaluation of energy efficiency, carbon emissions, and associated economic benefits;
\item Accuracy Analysis: A comparative evaluation of the compressed model’s performance relative to its full-size counterpart.

\end{enumerate}

\paragraph{}
We selected Llama 3.1 8B \cite{llama}, an LLM developed by Meta, as a baseline. This model was compared with a variant compressed by CompactifAI, we standardized calculation conditions and parameters to ensure experimental rigor.

\paragraph{}
To make comparable experiments, we ran all the tests on the same infrastructure which is a virtual machine from OVH cloud. We deployed our inference server on this machine and took measures to avoid any other tasks in parallel, so as not to interfere with our measurements. You can find below the characteristics of our inference server:

\begin{itemize}
    \item Infrastructure
        \begin{itemize}
            \item{Operating System: Linux (Kernel 5.15.0-130-generic, glibc 2.35)}
            \item{Processor: Intel® Xeon® Gold 6226R @ 2.90GHz}
            \item{GPU: 1 × NVIDIA Tesla V100S-PCIE-32GB}
            \item{CPU Cores: 15}
            \item{RAM: 43.05 GB}
            \item{Geographical Location: France (Latitude: 48.8582, Longitude: 2.3387)}
        \end{itemize}
    \item Software and Configurations
        \begin{itemize}
            \item{Python Version: 3.10.12}
            \item{AI Framework : PyTorch }
            \item{PyTorch Version: 2.5.1 }
            \item{CodeCarbon Version: 2.5.0}
            \item{CodeCarbon Tracking Mode: Machine-level monitoring}
        \end{itemize}
\end{itemize}

\subsubsection{Dataset}

\paragraph{}
We have compared our different models and configurations on one specific language task which is Question Answering. For all tests, we asked the system to answer 104 questions.
These questions were specifically designed to rigorously evaluate the performance across multiple domains :
\begin{itemize}
\item General Knowledge-Questions based on facts that incorporate historical occurrences, scientific facts, and widely accepted knowledge;
\item Scientific Inquiries- Complex questions based on a specific domain requiring expertise in physics, biology, chemistry, and engineering;
\item Location and Timestamp- Based Queries: Questions that rely on temporal and geographic context, ensuring the model can produce responses with real-world applicability;
\item Code and Process Generation- Tasks that assess the model’s capability to generate code snippets, debugging, and algorithmic explanations;
\item Context-Dependent Questions- Queries requiring reasoning based on previous information and inferred knowledge.
\end{itemize}

\paragraph{}
To ensure a balanced distribution of question length, complexity, and information dependencies, the dataset was carefully curated. Every query was structured to test the models' ability for reasoning, retrieval, and coherence. The diverse nature of the dataset enables a comprehensive evaluation of model compression's impact on different types of AI inference.

\subsubsection{Various Configurations}
\paragraph{}
Both models were evaluated with two distinct parameter configurations to analyze response quality across varied operational conditions:

\begin{itemize}

\item Balanced Response Generation (Set 1):

\begin{verbatim}
{
  "query": formatted_question,
  "sys_prompt": "You are a responsible AI. Provide text responses in JSON format.",
  "temperature": 0.5,
  "top_k": 50,
  "top_p": 0.5,
  "do_sample": "False",
  "min_new_tokens": 32,
  "max_new_tokens": 200
}
\end{verbatim}

\item Extended Response Generation (Set 2):

\begin{verbatim}
{
  "query": formatted_question,
  "sys_prompt": "You are a responsible AI. Provide text responses in JSON format.",
  "temperature": 0.1,
  "top_k": 50,
  "top_p": 0.5,
  "do_sample": "False",
  "min_new_tokens": 32,
  "max_new_tokens": 1000
}
\end{verbatim}

\end{itemize}
\paragraph{}
This analysis facilitates an assessment of the compressed model’s robustness across varying response generation constraints.

\subsection{Evaluation Metrics}

\subsubsection{Power consumption}

\paragraph{}
To evaluate the efficiency of compressed and full-size models we used CodeCarbon\cite{codecarbon}, an open-source tool designed to estimate the energy consumption and carbon footprint of executed code. It monitors the energy used by CPU, GPU and RAM and maps that energy to regional electricity grids to compute \ce{CO2} equivalent emissions. The main features of CodeCarbon are :

\begin{itemize}
\item Hardware Monitoring – CodeCarbon recognizes CPU and GPU specifications and tracks power consumption in real time;
\item Power estimation – CodeCarbon uses proxies to estimate the energy consumption of CPU (when it can't be directly tracked with RAPL) and RAM by using their specifications and a predefined mapping of hardware energy draw (Watt-hour estimates per component);
\item \ce{CO2} Estimation – It converts the total energy consumption into kg \ce{CO2} equivalent while taking into consideration the local energy mix based on the geographical location of the hardware;
\item Logging \& Reporting – It helps us to store the energy information for analysis, to compare models and optimizations for sustainability.
\end{itemize}

\paragraph{}
In our research, CodeCarbon was integrated into the inference pipeline. The tracker was initialized before model execution and it was terminated after response generation. 
Experimental conditions were carefully controlled, ensuring precision in measurement through consistent  hardware configurations.

\subsubsection{Accuracy}\label{methodo-accuracy}

\paragraph{}
To evaluate the performance of models, we used Ragas\cite{ragas}, a standardized AI benchmarking framework that provides customizable evaluation metrics for LLM applications. These metrics are specially designed to measure different aspects of system performance. It provides both traditional non-LLM-based metrics, which focus on retrieval precision, token overlap, and similarity scores, as well as LLM-based metrics that use language model reasoning to assess aspects like factual correctness, semantic alignment, and contextual relevance. Ragas is able to deliver both objective and interpretive insights due to its hybrid nature. Notably, each metric is modular and can be modified to suit application requirements, providing exceptional flexibility. This makes Ragas particularly useful for evaluating LLMs, where dynamic interactions and retrieval accuracy are important. By incorporating both fundamental and nuanced evaluations, Ragas facilitates comprehensive benchmarking and ensures that LLM applications are assessed in a robust and insightful manner.

\paragraph{}
To perform these accuracy tests, we used the model ChatGTP 4o as a reference, to generate the ground truth responses to the questions of the dataset. Then, we also generated the responses with the compressed and uncompressed models and calculated the following Ragas Metrics for LLM evaluation:
\begin{itemize}
\item ROUGE Score [0-1] – It measures the degree of lexical overlap between generated responses and reference texts (ground truth);
\item BLEU Score [0-1] – It evaluates syntactic integrity and fluency;
\item Semantic Similarity [0-1] – It uses a bi-encoder model to calculate the degree to which the generated response aligns with the ground truth in meaning;
\item Factual Correctness [0-1] – It measures how accurately a generated response aligns with a reference by breaking it into claims and using natural language inference. It utilizes precision, recall, and F1 scores, as well as allowing control  over atomicity (degree of claim decomposition) and coverage (extent of information retained);
\item Answer Correctness Score [0-1] – It evaluates how accurately a model's generated answers match the ground truth answers using similarity measures. It provides a score indicating the correctness of responses, which is crucial for assessing model performance in NLP tasks;
\item Response Relevancy Score [0-1] – It compares embeddings through cosine similarity to measure how well a generated answer aligns with the user query. It ensures that the answer directly addresses the original question by penalizing incomplete or redundant responses.
\end{itemize}

\paragraph{}
All these scores range from 0 to 1, with higher values indicating better semantic alignment.

\section{Results}
\subsection{Power Consumption}

\paragraph{}
In the next subsections, we will share the figures we were able to measure (with CodeCarbon) during our 4 tests (compressed model \& full-size with 2 configurations each) to complete our Q\&A task on all 104 questions.

\subsubsection{Total Duration (in minutes):}

\begin{itemize}
    \item The compressed model required 10.29 min to complete the task for 200-tokens configuration and 23.56 min for 1000-tokens configuration;
    \item The full-size model required 10.91 min for 200-tokens configuration and 28.79 min for 1000-tokens configuration.
\end{itemize}
Both tests on the compressed model show a reduction in processing time, (-5.68\%) for 200 token configuration and a more significant reduction in the 1000 token configuration (-18.17\%).

\subsubsection{Total greenhouse gas emissions (in kg \ce{CO2} eq.):}

\begin{itemize}
    \item The compressed model emitted 2.75E-03 kg \ce{CO2} eq. for 200-tokens configuration and 6.40E-03 kg \ce{CO2} eq. for 1000-tokens configuration;
    \item The full-size model emitted 3.93E-03 kg \ce{CO2} eq. for 200-tokens configuration and 1.05E-02 kg \ce{CO2} eq. for 1000-tokens configuration.
\end{itemize}
\paragraph{}
We can notice that \ce{CO2} eq. emissions are lower in case of compressed models by approximately 30.03\% for 200-token responses and 39.05\% for 1000-token responses.

\paragraph{}
All tests were run on the same virtual machine located at the same place in France, a carbon intensity of 5.6039E-02 kg \ce{CO2} eq./kWh have been applied.

\begin{figure}[h]
    \centering
    \includegraphics[width=.69\textwidth]{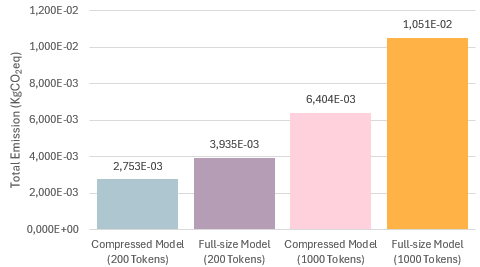}
    \caption{Total carbon emissions across tests}
    \label{fig:carbon}
\end{figure}

\FloatBarrier

\subsubsection{Energy Consumption}

\paragraph{Total CPU Energy:}

\begin{itemize}
    \item The compressed model consumed 1.29E-02 kWh for 200-tokens configuration and 2.94E-02 kWh for 1000-tokens configuration;
    \item The full-size model consumed 1.36E-02 kWh for 200-tokens configuration and 3.60E-02 kWh for 1000-tokens configuration.
\end{itemize}

\paragraph{}
The compressed model consumes less CPU energy ~5.79\% (200-tokens), ~18.15\% (1000-tokens).

\paragraph{Total GPU Energy:}

\begin{itemize}
    \item The compressed model consumed 1.68E-02 kWh for 200-tokens configuration and 3.97E-02 kWh for 1000-tokens configuration;
    \item The full-size model consumed 2.98E-02 kWh for 200-tokens configuration and 8.01E-02 kWh for 1000-tokens configuration.
\end{itemize}

\paragraph{}
The compressed model consumes less GPU energy ~43.55\% (200-tokens), ~50.5\% (1000-tokens).

\paragraph{Total Energy Consumed:}

\begin{figure}[h]
    \centering
    \includegraphics[width=.69\textwidth]{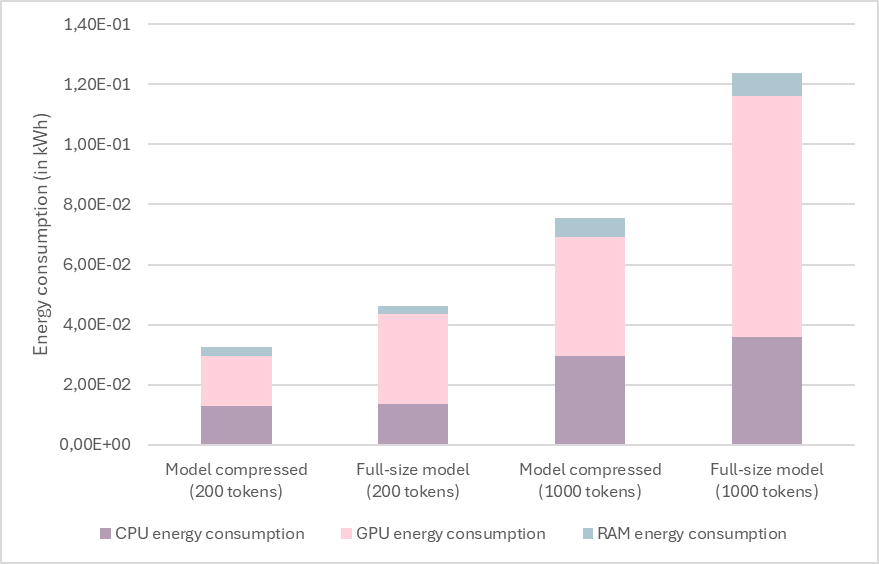}
    \caption{Total energy consumption broken down by component across tests}
    \label{fig:rep_energy_1}
\end{figure}

\begin{figure}[h]
    \centering
    \includegraphics[width=.69\textwidth]{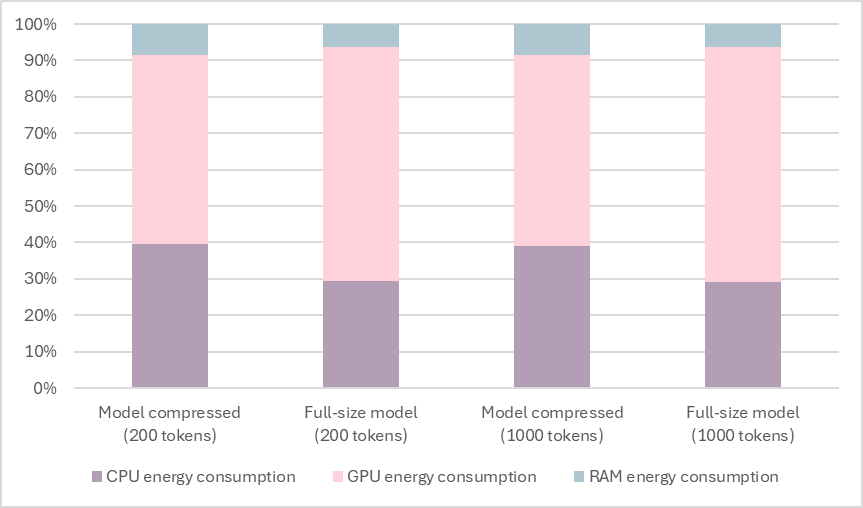}
    \caption{Dispatch of energy consumption broken down by component across tests}
    \label{fig:rep_energy_2}
\end{figure}

\paragraph{}
We can see on the Figure \ref{fig:carbon}  \ref{fig:rep_energy_1}  that the test with compressed model consumes less energy that the full-size ones, for both configuration. In particular we notice 30.04\% reduction for the 200-tokens configuration and 39.09\% for the other one. We also note (Figure \ref{fig:rep_energy_2}) that most of the power consumption comes from the GPU, while that of RAM is almost negligible. Finally, we can also see that the compressed model uses less GPU than the full-size one.

\FloatBarrier

\paragraph{Analysis of parameters influencing consumption}
    
\paragraph{}
We wanted to dig deeper to better understand the parameters that influence consumption. We noticed that the server's energy consumption is very correlated with the number of token generated (98.65\%), whereas the number of tokens in the prompts has no significative direct correlation with the energy consumption. However, we carried out a more in-depth analysis, because in the test dataset we had 10 questions asked without a context, and the same 10 questions with a context in which the agent should retrieve the answers. The increase of energy consumption for the questions with context is between 0.4 and 3\%. 

\paragraph{}
In the following, we will focus on the relationship between the number of tokens generated and the energy consumed. We performed a linear regression to establish a coefficient between the number of token generated and the energy consumption. 

\begin{figure}[h]
    \centering
    \includegraphics[width=.69\textwidth]{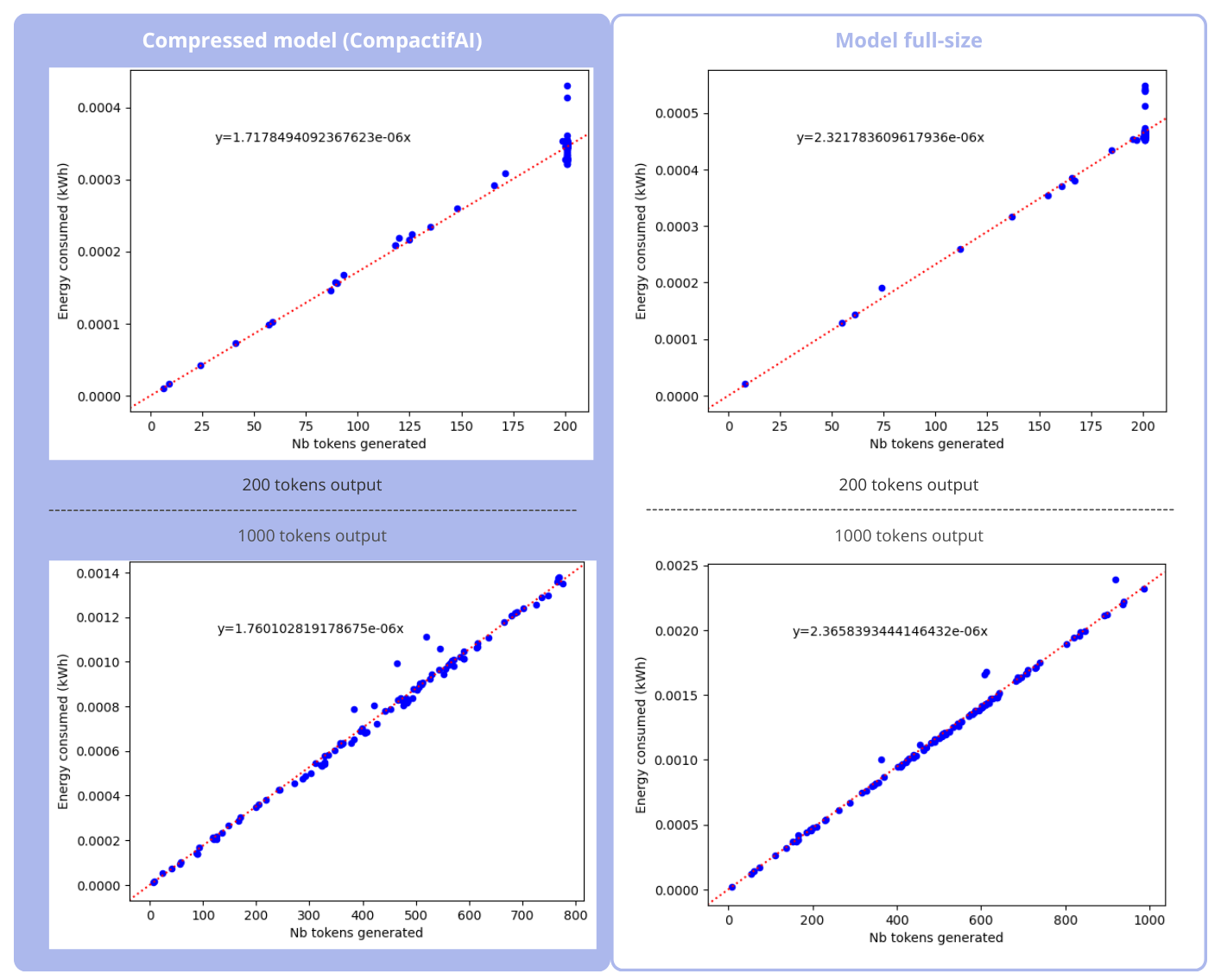}
    \caption{Linear regression curve and experimental points for each of the 4 tests}
    \label{fig:slope_graphs}
\end{figure}

\begin{figure}[h]
    \centering
    \includegraphics[width=.69\textwidth]{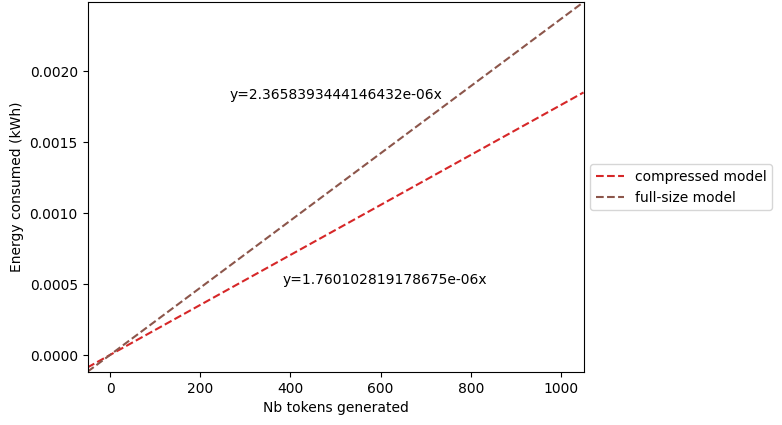}
    \caption{Comparison of the 2 models in terms of consumption per token generated}
    \label{fig:comp_per_token}
\end{figure}

\paragraph{}
We can see the Figure \ref{fig:slope_graphs} that linear regression fits the situation, since the points follow the curve perfectly. This is a little less true in the top graphs with the output limit of 200 tokens, since the limit was too low and most of the responses were cut off in their generation, which produces a lot of points for 200 tokens, with a large variance. If we compare only the two graphs below with the output limit at 1000 tokens, we obtain the Figure \ref{fig:comp_per_token}.

\paragraph{}
We can see that the compressed model consumes -25.60\% energy to generate one token. 
Furthermore, we have noticed that on average the compressed model generates 18,11\% fewer tokens (i.e. it answers questions more succinctly). If we apply this two diminutions together, we obtain a reduction of 39,08\% which is very close from what we observed in the previous part \textit{Total Energy Consumed} (39,09\%).

\paragraph{}
According to the analysis, the compressed model significantly reduce computational overhead and require less processing time than its full-size version. It can be observed that energy efficiency in GPU power usage is most prominent, where compressed models demonstrate approximately 40-50\% lower GPU power consumption. The total emissions are reduced by ~30-39\%, which highlights the sustainability benefits of model compression. Also, the reduction in power consumption and carbon emissions is more pronounced for longer responses (1000-token configuration), reinforcing the effectiveness of compression in extended tasks. It leads to reduction in hardware strain and financial savings, making AI model deployment more sustainable and cost-effective.

\FloatBarrier

\subsection{Accuracy Performance}

\paragraph{}
To guarantee the performance and therefore the usability of the compression method, we also decided to test it from the point of view of the quality of the responses generated. To do this, we calculated various accuracy metrics, frequently used to assess performance in Q\&A tasks. Following the methodology described in the section \ref{methodo-accuracy}, we obtained the following results.

\subsubsection{ROUGE Score (Lexical Overlap)}

\begin{itemize}
    \item ROUGE (200 tokens): Compressed Model (0.2545) \textgreater  Full-size Model (0.2471).
    \item ROUGE (1000 tokens): Compressed Model (0.2022) \textgreater  Full-size Model (0.1835).
\end{itemize}

\begin{figure}[h]
    \centering
    \includegraphics[width=.69\textwidth]{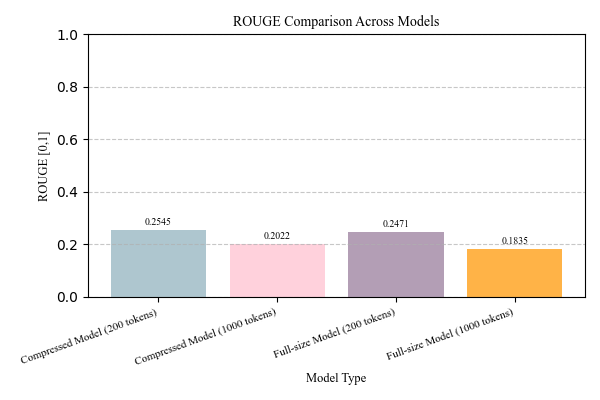}
    \caption{Comparison of ROUGE Score across tests}
    \label{fig:rouge}
\end{figure}

\hfill

\paragraph{}
The compressed model perform better compared to full-size models. These results imply that compressed models maintain higher lexical fidelity to ground truth than their full-size counterparts. See Figure \ref{fig:rouge}.

\FloatBarrier

\subsubsection{BLEU Score (Grammatical Integrity and Fluency)}

\begin{itemize}
    \item BLEU (200 tokens): Compressed Model (0.1776) \textgreater  Full-size Model (0.1321)
    \item BLEU (1000 tokens): Compressed Model (0.1774) \textgreater  Full-size Model (0.1319)
\end{itemize}

\begin{figure}[h]
    \centering
    \includegraphics[width=.69\textwidth]{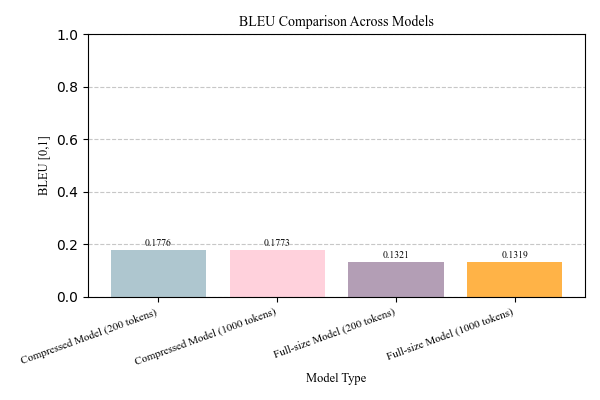}
    \caption{Comparison of BLEU Score across tests}
    \label{fig:bleu}
\end{figure}

\paragraph{}
The compressed model exhibit improved syntactic coherence.These results indicate that the compressed model produce better responses with grammatical precision and improved structural fluency. See Figure \ref{fig:bleu}.

\FloatBarrier

\subsubsection{Semantic Similarity (Contextual Fidelity)}

\begin{itemize}
    \item Semantic Similarity (200 tokens): Compressed Model (0.7775) \textgreater  Full-size Model (0.7752)
    \item Semantic Similarity (1000 tokens): Compressed Model (0.7627) \textgreater  Full-size Model (0.7514)
\end{itemize}

\begin{figure}[h]
    \centering
    \includegraphics[width=0.69\textwidth]{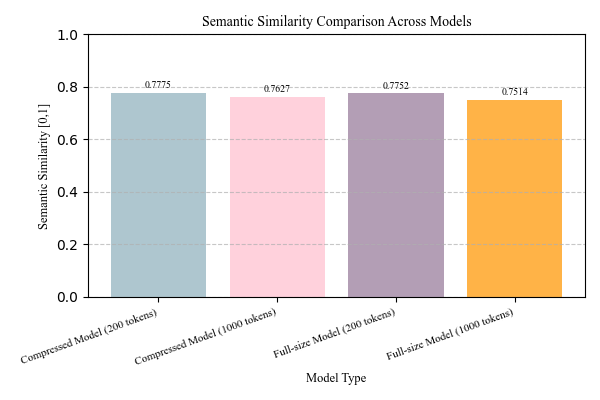}
    \caption{Comparison of Semantic Similarity Score across tests}
    \label{fig:semantic}
\end{figure}

\paragraph{}
There was a negligible variance observed across compression conditions. Even with the size reduction of the model, semantic coherence remains stable. See Figure \ref{fig:semantic}.

\FloatBarrier

\subsubsection{Factual Correctness}

\begin{itemize}
    \item Factual Correctness (200 tokens): Full-size Model (0.6072) \textgreater  Compressed Model (0.5732)
    \item Factual Correctness (1000 tokens): Full-size Model (0.5873) \textgreater  Compressed Model (0.5845)
\end{itemize}

\begin{figure}[h]
    \centering
    \includegraphics[width=0.69\textwidth]{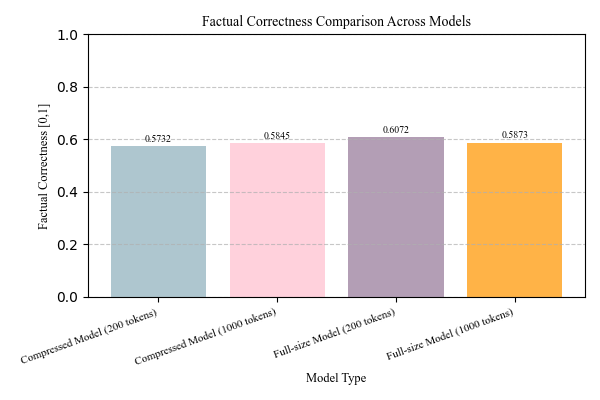}
    \caption{Comparison of Factual Correctness Score across tests}
    \label{fig:factual}
\end{figure}

\paragraph{}
It can be observed that the full-size model exhibits a marginal advantage over the compressed model.

\paragraph{}
We also calculated Factual correctness exclusively for context-based questions due to the way this metric operates. Factual correctness evaluates the truthfulness and factual grounding of the output, in contrast to answer correctness and response relevancy, which assess the semantic alignment and intent fulfillment of the response. This is accomplished by decomposing both the generated response and the reference into individual claims, followed by natural language inference to determine factual overlap. Context-based queries particularly require such granular and inference-driven evaluation as the risk of hallucination is higher. Hence, factual correctness was selectively applied to ensure a precise and claim-level verification of factual alignment, making it highly suitable for validating answers based on the provided context. See Figure \ref{fig:factual}.

\FloatBarrier

\subsubsection{Answer Correctness Score}

\begin{itemize}
    \item Answer Correctness Score (200 tokens): Compressed Model (200 tokens) (0.5579) \textgreater  Full-size Model (200 tokens) (0.5530)
    \item Answer Correctness Score (1000 tokens): Compressed Model (1000 tokens) (0.5876) \textgreater  Full-size Model (1000 tokens) (0.5714)
\end{itemize}

\begin{figure}[h]
    \centering
    \includegraphics[width=0.69\textwidth]{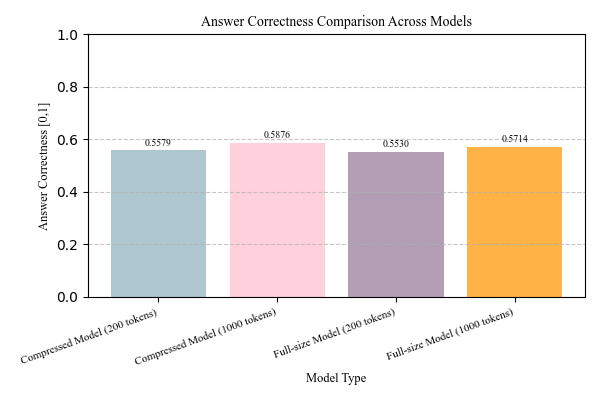}
    \caption{Comparison of Answer Correctness Score across tests}
    \label{fig:correctness}
\end{figure}

\paragraph{}
According to the score the compressed model perform slightly better. Accuracy is not substantially reduced by compression. See Figure \ref{fig:correctness}.

\FloatBarrier

\subsubsection{Response Relevancy Score}

\begin{itemize}
    \item Response Relevancy Score (200 tokens): Full-size Model (200 tokens) (0.7224) \textgreater  Compressed Model (200 tokens) (0.7152)
    \item Response Relevancy Score (1000 tokens): Full-size Model (1000 tokens) (0.7244) \textgreater  Compressed Model (1000 tokens) (0.7051)
\end{itemize}

\begin{figure}[h]
    \centering
    \includegraphics[width=0.69\textwidth]{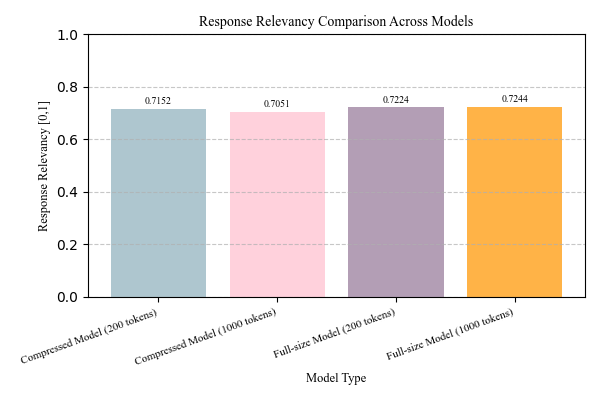}
    \caption{Comparison of Response Relevancy Score across tests}
    \label{fig:resrel}
\end{figure}

\paragraph{}
The full-size model handle longer contextual dependencies more effectively, preserving better alignment with user queries but the difference is not 
significant in comparison to compressed models.

\paragraph{}
According to the analysis of accuracy matrices it can be observed that compressed models demonstrate superior performance in ROUGE and BLEU metrics, indicating effective lexical and grammatical coherence compared to full-size version. Semantic similarity remains largely unaffected by compression ensuring that contextual alignment is maintained. However, the full-size model retain a marginal advantage in factual correctness and response relevancy, indicating a minor compromise in information precision. Notably, the compressed model successfully achieve almost equivalent accuracy while significantly reducing computational overhead.

\paragraph{}
Empirical results show that model compression yields high level of computational efficiency while preserving a high degree of accuracy fidelity. See Figure \ref{fig:resrel}.

\FloatBarrier

\section{Discussions}
In this section we wanted to focus on the limitations we have identified about our analysis.

\begin{itemize}
\item We were not able to compare the results especially power consumptions with other framework than PyTorch.  
\textit{For instance, by experience VLLM framework seems to be more efficient.}

\item We were not able to test other compressed models to verify that the results are the same and do not depend on the model. 
\textit{For example : Multiverse Computing says that "the bigger the model, the bigger the benefits of the compression in terms of energy saving and accuracy performance."}

\item We did not compare with other methods of compression like open source ones (For example : quantization)
\textit{By experience we have noted gains in power consumption (30\%) from quantization but with significant loss of accuracy}

\item We do not have information about the impacts of the compression itself. 
\textit{This is important because that would enable one to calculate how long (or how many usages) it would take to amortize the compression and when the overall global impact would be positive.}
\end{itemize}

\paragraph{}
Of course their might be more limitations that we overlooked but we think highlighted here the most important ones.

\section{Conclusion}
\paragraph{}
The research confirms that the model compressed via CompactifAI provide substantial reductions in power consumption, operational costs, and carbon emissions while maintaining competitive accuracy. These results demonstrate the potential of AI compression techniques in addressing the challenges related to excessive computational, financial, and ecological impacts associated with large-scale AI deployment.

\FloatBarrier

{
       \bibliographystyle{unsrt}

\end{document}